\documentclass[10pt,twocolumn,letterpaper]{article}

 \usepackage{iccv}              
\usepackage{multirow} 
\definecolor{iccvblue}{rgb}{0.21,0.49,0.74}
\usepackage[pagebackref,breaklinks,colorlinks,allcolors=iccvblue]{hyperref}

\title{DiffV2IR: Visible-to-Infrared Diffusion Model via  Vision-Language Understanding}

\author{Lingyan Ran \\
Northwestern Polytechnical University\\
\and
Lidong Wang\\
Northwestern Polytechnical University\\
\and
Guangcong Wang\\
Great Bay University\\
\and
Peng Wang\\
Northwestern Polytechnical University
\and
Yanning Zhang\\
Northwestern Polytechnical University
}

\begin{document}
\maketitle

\begin{abstract}

The task of translating visible-to-infrared images (V2IR) is inherently challenging due to three main obstacles: 1) achieving semantic-aware translation, 2) managing the diverse wavelength spectrum in infrared imagery, and 3) the scarcity of comprehensive infrared datasets. Current leading methods tend to treat V2IR as a conventional image-to-image synthesis challenge, often overlooking these specific issues. To address this, we introduce DiffV2IR, a novel framework for image translation comprising two key elements: a Progressive Learning Module (PLM) and a Vision-Language Understanding Module (VLUM). PLM features an adaptive diffusion model architecture that leverages multi-stage knowledge learning to infrared transition from full-range to target wavelength. 
To improve V2IR translation, VLUM incorporates unified Vision-Language Understanding. 
We also collected a large infrared dataset, IR-500K, which includes 500,000 infrared images compiled by various scenes and objects under various environmental conditions. Through the combination of PLM, VLUM, and the extensive IR-500K dataset, DiffV2IR markedly improves the performance of V2IR. Experiments validate DiffV2IR's excellence in producing high-quality translations, establishing its efficacy and broad applicability. The code, dataset, and DiffV2IR model will be available at \url{https://github.com/LidongWang-26/DiffV2IR}.

\end{abstract}

\section{Introduction}

\begin{figure}[htbp]
    \centering
    \includegraphics[width=0.95\linewidth]{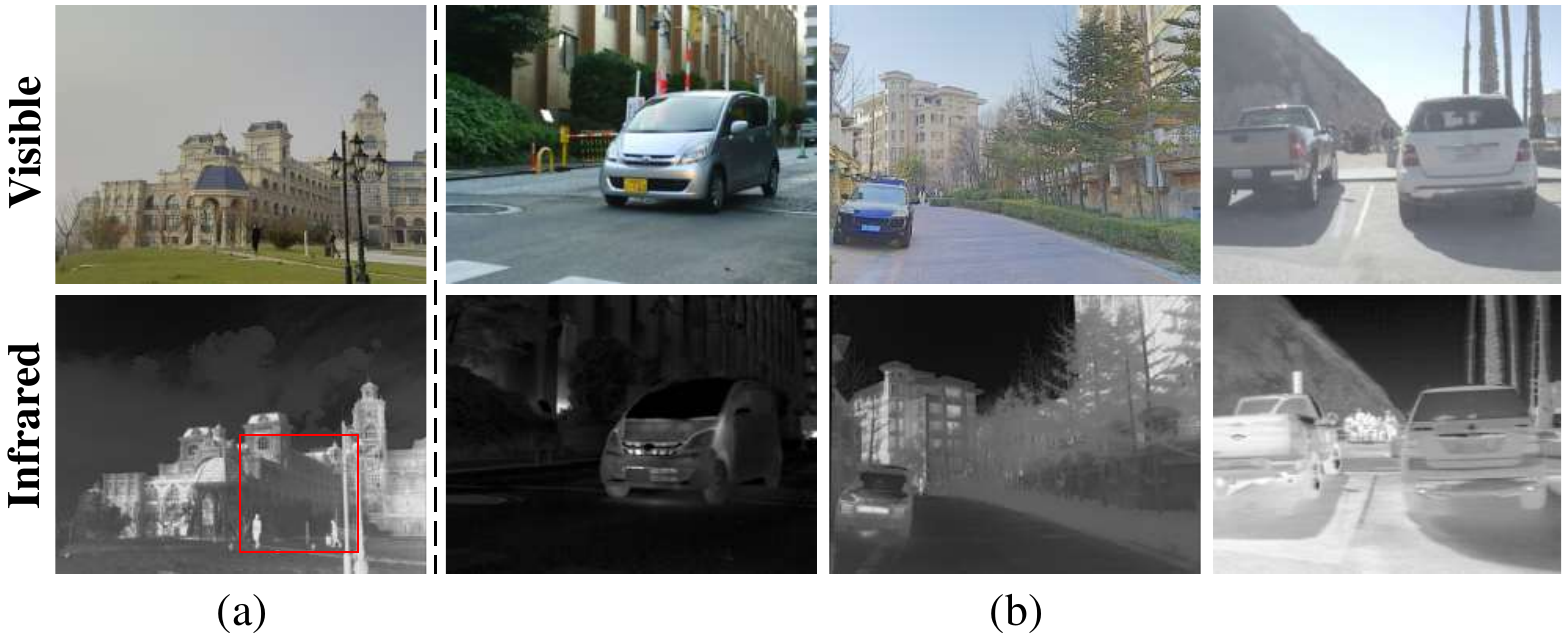}
    \caption{Main challenges of V2IR. (a) Semantic-aware translation, in which the context information of shadow influences the infrared image a lot. (b) Diverse infrared radiations. Even similar visual scenes from different infrared cameras show the diversity of infrared imagery. For the second column to the fourth column, the infrared intensity significantly changes.}
    \label{fig:motiv}
\end{figure}

Visible images are the most common form of digital imagery, while infrared images hold significant importance in various practical applications, such as thermal imaging, night vision monitoring, and environmental sensor data analysis. Currently, lots of foundation visual models are built on annotated large visible datasets. An intuitive way to narrow the gap between visual models and infrared models is to translate visual images into infrared images (V2IR).


However, V2IR presents a difficult task due to three critical challenges. \textit{\textbf{First}}, V2IR is highly dependent on semantic information. Infrared imaging, also known as thermal imaging, captures and visualizes the infrared radiation emitted by objects. Therefore, V2IR relies on scene understanding such as object semantics and context information (e.g., solar radiation, lighting, and shadow). \textit{\textbf{Second}}, infrared imaging primarily relies on infrared radiation, which is typically divided into several different wavelength ranges such as Near-Infrared (NIR), Short-Wave Infrared (SWIR), Mid-Wave Infrared (MWIR), and Long-Wave Infrared (LWIR). Different infrared radiation intensity leads to various pixel values for the same scene. Images captured in the same wavelength range might still vary due to differences in infrared camera sensors. \textit{\textbf{Third,}} unlike visual images that can be easily captured by anyone with widely-used phones and traditional cameras, infrared images are limited to a few people who have infrared cameras. Therefore, there are only a few public infrared datasets captured by different types of infrared camera. It is unclear how to train a large foundation model of infrared image generation on limited datasets.


Existing methods make initial attempts at V2IR. A common way is to directly formulate the V2IR task as image-to-image translation, with methods such as Variational Autoencoders (VAEs)~\cite{vae,rezende2014stochastic,hwang2020variational}, Generative Adversarial Networks (GANs)~\cite{wgan,isola2017image,Lin_2018_CVPR}, and diffusion models~\cite{dm,ddpm,ldm}. 
To exploit semantic information (Challenge 1) and reduce the impact of diverse infrared radiation from different infrared cameras (Challenge 2), these methods integrate different low-level semantics into image-to-image translation models, such as edge prior, structural similarity, geometry information, and physical constraint. 
For example, InfraGAN~\cite{infragan} uses structural similarity as an additional loss function and a pixel-level discriminator. EG-GAN~\cite{eggan,unit} focused on edge preservation. DR-AVIT~\cite{han2024dr} achieved diverse and realistic aerial visible-to-infrared image translation by integrating a geometry-consistency constraint. TMGAN~\cite{matchgan} incorporates the image-matching process into image-to-image translation. PID~\cite{mao2024pid} incorporated strong physical constraints and used a latent diffusion model. Although these methods significantly improve V2IR, they do not make full use of semantics and do not consider different infrared radiation from different infrared cameras.

To tackle the challenges of V2IR, in this paper, we present DiffV2IR, a novel V2IR diffusion framework that integrates vision-language understanding into a diffusion model with multi-stage knowledge learning. Specifically, to achieve semantic-aware V2IR translation, we extract a detailed scene description by integrating a Vision-Language Understanding Module (VLUM) into the optimization (Challenge 1). To achieve stable V2IR translation trained on the datasets that contain different infrared radiation from different infrared cameras, we propose a Progressive Learning Module (PLM) that features an adaptive diffusion model architecture that leverages multi-stage knowledge learning to transition from full-range to target wavelength (Challenge 2). To train a large-scale V2IR diffusion model, we assembled an extensive infrared dataset named IR-500K, comprising 500,000 infrared images. The IR-500K integrates nearly every substantial publicly accessible infrared dataset. This fusion of scale, diversity, and accessibility establishes the dataset as a pivotal resource for enhancing infrared image generation technologies(Challenge 3). 
With DiffV2IR and IR-500K, our work significantly improves the performance of V2IR.

Overall, the main contributions are: 1) We propose a novel DiffV2IR framework that integrates a multi-modal vision-language model into a unified optimization and thus achieves semantic-aware V2IR translation. 2) To enable DiffV2IR to perform stable V2IR translation on an infrared dataset with various infrared radiation, we propose a progressive learning module that leverages multi-stage knowledge learning. 3) To train a large DiffV2IR model, we collect a large infrared dataset, IR-500K. Experiments demonstrate the effectiveness of DiffV2IR.

\section{Related Works}

\textbf{Diffusion Models.} 
Recently, diffusion models such as those in~\cite{yang2023diffusion,DDIM,improvedddpm} have made significant strides in image generation. Acting as generative systems, these models emulate physical diffusion processes by gradually introducing noise to learn how to produce clear images. Unlike traditional generative models like GANs~\cite{goodfellow2014Generative,liu2016coupled,mao2017least} and VAEs~\cite{vae,rezende2014stochastic}, diffusion models generate superior-quality samples for high-resolution image creation and provide a training process less prone to mode collapse. The concept of diffusion models was first presented in~\cite{dm}. DDPMs \cite{ddpm} proposed  denoising diffusion probabilistic models, which captured significant attention and broadened the application of diffusion models in image generation. Efforts have since been aimed at enhancing their efficiency and production quality. Latent diffusion models (LDMs)~\cite{ldm} executed the diffusion process in a compressed latent space, greatly reducing computational overhead. These models have excelled in image generation and denoising. However,  they remain underutilized in multispectral image translation.

\textbf{Image-to-Image Translation.} Image translation algorithms are designed to learn either a pixel-wise correspondence or a joint probability distribution to facilitate the translation of images from one domain to another. 
Pix2Pix~\cite{pix2pix}, a foundational work in the field, utilizes a conditional generative adversarial network (cGAN)~\cite{cGAN} to develop a pixel-level map between input and output images. 
Expanding on this, Pix2PixHD~\cite{Wang_2018_CVPR} explores techniques for producing high-resolution images of superior quality. These methods require paired images for training. 
Introducing a different approach, CycleGAN~\cite{cyclegan}, DiscoGAN~\cite{discogan} and DualGAN~\cite{dualgan} utilizes unpaired datasets by implementing a cycle consistency loss, which guarantees that the mapping from source to target and back to source retains the original content. 
Then many models like~\cite{choi2018stargan,choi2020stargan,baek2021rethinking,xie2021unaligned,Wu_2024_CVPR} ultilize cycle consistency for unpaired training. ~\cite{lee2018diverse,huang2018multimodal,Usman_2023_WACV} assume that the representation can be disentangled into domain-invariant semantic structure features and domain-specific style features. ~\cite{alami2018unsupervised,ugatit,tang2021attentiongan,kim2022instaformer} implement attention mechanism in image translation.


With the rapid advancement of diffusion models, a variety of conditional diffusion models that incorporate text and spatial information have achieved notable success in image translation~\cite{tumanyan2023plug,xia2024diffi2i,fcdiffusion,styleid,Cho_2024_CVPR,lee2024diffusion}. InstructPix2Pix~\cite{ip2p} employs two large pre-trained models (GPT-3 and Stable Diffusion) to generate an extensive dataset of input-goal-instruction triplet examples and trains a model for image editing based on instructions using this dataset. ControlNet~\cite{controlnet} and T2I-Adapter~\cite{t2i} are devoted to making the diffusion process more controllable by introducing various conditions. 
Although many methods achieve great success in image translation, they do not consider the challenges of V2IR, which heavily depends on scene understanding.


\begin{figure*}[htbp]
    \centering
    \includegraphics[width=0.9\linewidth]{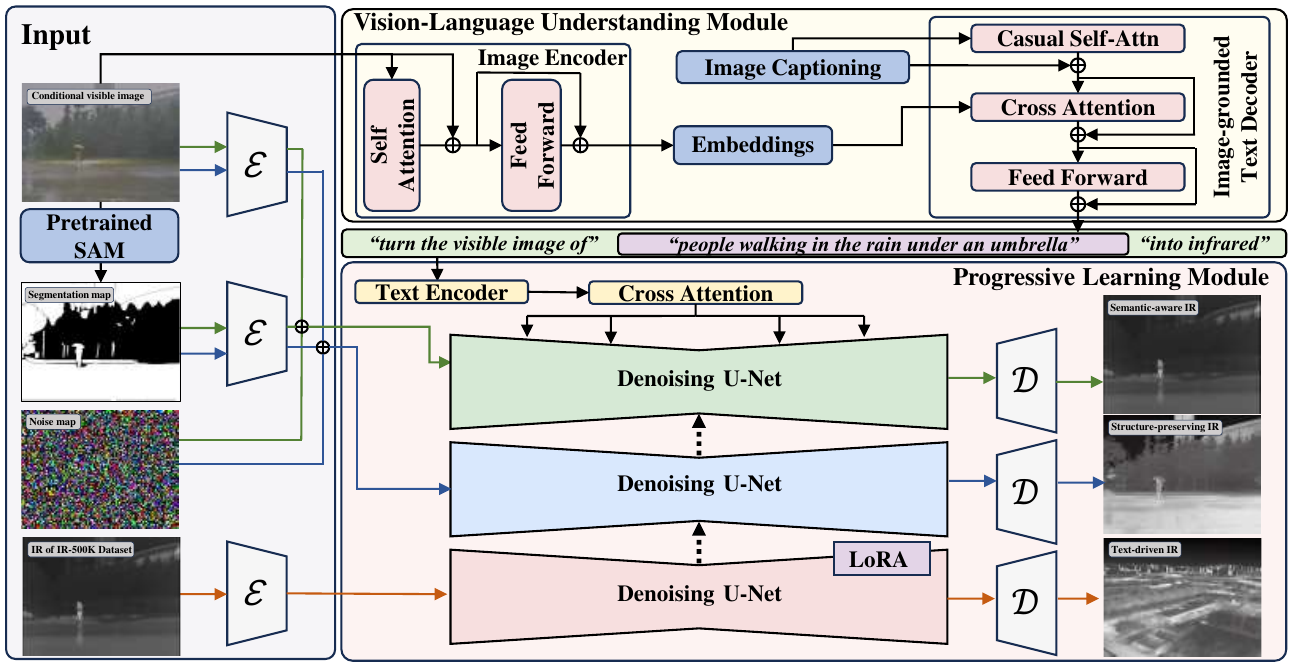}
    \caption{Framework overview of our DiffV2IR. DiffV2IR mainly consists of two components, i.e., Progressive Learning Module (PLM) and Vision-Language Understanding Module (VLUM). We use PLM for multi-stage knowledge learning and VLUM for semantic preserving in the V2IR task. The three U-Nets from bottom to top respectively denote the infrared representation internalization phase, the cross-modal transformation learning phase, and stylization refinement phase of PLM. The VLUM is introduced during PLM to make DiffV2IR semantic-aware.}
    \label{fig:framework}
\end{figure*}

\textbf{Visible-to-Infrared Image Translation.} 
Several models have attempted to translate visible images to infrared images. Initially, some research focused on generating infrared data tailored for specific tasks like tracking~\cite{zhang2018synthetic} and person re-identification~\cite{kniaz2018thermalgan}, treating it mainly as a pixel generation challenge. Content structure serves as a crucial prior in producing meaningful infrared images. InfraGAN~\cite{infragan} incorporates structural similarity as an auxiliary loss and uses a pixel-level discriminator for V2IR image translation. EG-GAN~\cite{eggan,unit} highlights edge preservation as an effective approach, confirmed by improved outcomes in training deep TIR optical flow and object detection against other benchmarks. VQ-InfraTrans~\cite{sun2023vq} introduces a two-step transfer strategy using a composite encoder and decoder from VQ-GAN~\cite{vq-gan}, alongside a multi-path transformer. DR-AVIT~\cite{han2024dr} enhances the translation of aerial visible-to-infrared images with disengaged representation learning, separating image representations into a domain-invariant semantic structure space and two domain-specific imaging style spaces. PID~\cite{mao2024pid} further advances this area by integrating significant physical constraints and for the first time employing a latent diffusion model. However, current V2IR methods do not fully exploit key semantic information and struggle to create high-quality infrared images since various infrared imaging.




\section{Methodology}
In this paper, we propose a DiffV2IR framework based on diffusion models, which can translate visible images into infrared images (V2IR). Different from large-scale visual image datasets that can easily collected by widely-used phones and RGB cameras, it is difficult to collect diverse visual images in various scenes with limited infrared cameras. To address this problem, we collect a large-scale dataset by combining almost all publicly available infrared datasets (Section \ref{sec:dataset}). However, different infrared cameras might process diverse wavelength spectrums, leading to various infrared images. We design a progressive learning method to learn multi-stage infrared knowledge (Section \ref{sec:pl}). Since the V2IR translation is highly dependent on semantic information (object semantics and context information), we design a semantic-aware V2IR translation module via vision-language understanding (Section \ref{sec:semantics}).



\subsection{Preliminary}
\label{sec:pre}
Diffusion models are a family of probabilistic generative models that progressively destruct data by injecting noise,  then learn to reverse this process for sample generation. 
DDPMs~\cite{ddpm} are probabilistic generative model leveraging two Markov chains. The first Markov chain progressively injects noise into the data to transform  data distribution into standard Gaussian distribution, while the other stepwise reverses the process of noise injection, generating data samples from Gaussian noise. 
LDMs~\cite{ldm} significantly reduce resource demand by operating in the latent space, especially dealing with high-resolution images. 
LDMs mainly consists of an autoencoder with an encoder $\mathcal{E}$ and a decoder $\mathcal{D}$ and a denoising U-Net $\epsilon_{\theta}$. 
Given an image $x$, LDMs first encode it into latent space and then add noise to the encoded latent $z=\mathcal{E}(x)$ producing a noisy latent $z_{t}$, where $t$ denotes diffusing time step. For conditional diffusion models, condition $c$ is introduced into the denoising process. The denoising U-Net $\epsilon_{\theta}$ is trained by minimizing the following objective:
\begin{equation}
\begin{split}
L = \mathbb{E}_{\mathcal{E}(x), c, \epsilon \sim \mathcal{N}(0,1), t} \left[ \left\| \epsilon - \epsilon_{\theta} \left( z_t, t, c \right) \right\|_2^2  \right], 
\end{split}
\end{equation}
where LDMs aim to predict the noise added on the encoded latent $z$ at $t$ timesteps under the condition $c$.

\subsection{Overview of DiffV2IR }
\label{sec:overview}
The pipeline of our DiffV2IR is illustrated in Figure \ref{fig:framework}. 
DiffV2IR mainly consists of two components, i.e., Progressive Learning Module (PLM) and Vision-Language Understanding Module (VLUM). 
Specifically, 1) as for PLM, we first establish foundational knowledge of infrared imaging properties utilizing our collected IR-500K dataset. 
Then we use visible-infrared image pairs to learn cross-modal transformation and finally conduct the refinement on the specific infrared imaging style. 
2) as for VLUM, we incorporate unified vision-language understanding, including detailed language descriptions and segmentation maps, to make DiffV2IR semantic-aware and structure-preserving. 

\subsection{Visible-to-Infrared Diffusion Model via Progressive Learning }
\label{sec:pl}

We employ the IR-500K Dataset, as described in Section \ref{sec:dataset}, to train a conditional diffusion model aimed at converting visible images into infrared ones. Typically, fine-tuning diffusion models from a pre-trained checkpoint yields better results than training a diffusion model from scratch. Consequently, our model is constructed upon Stable Diffusion (SD), a pre-trained latent diffusion model with text conditioning~\cite{ldm}, to leverage its extensive expertise in the domain of text-to-image translation.

Although pre-trained diffusion models are capable of generating high-quality visible images based on textual prompts, we observed that their performance degrades when the prompt includes the term ``\textit{infrared}." This indicates a generally poor comprehension of infrared modality for most pre-trained diffusion models, with even less proficiency in translating visible images to infrared ones.
To address this, we introduce PLM, a progressive learning strategy. This approach firstly enables the diffusion model to fill the gap between infrared modality and visual modality, then develops its capability of performing general visible-to-infrared image translation, and finally allows it to generate infrared images in a specified style.

\textbf{Phase \#1: infrared representation internalization,} which aims to establish foundational knowledge of infrared imaging properties.
This is the initial phase in our progressive learning strategy designed to integrate infrared knowledge. We achieve this by fine-tuning a stable diffusion model using Low-Rank Adaptation (LoRA)~\cite{hu2022lora} on IR-500K dataset, all prompted with the same phrase, \textit{``an infrared image"}. Throughout this tuning process, the weights of the pre-trained model remain fixed, while smaller trainable rank decomposition matrices are inserted into the model, enhancing training efficiency and minimizing overfitting risks. As a result of this progressive learning phase, the diffusion model associates infrared characteristics with the term ``\textit{infrared}", enabling the generation of infrared-style images from textual prompts, without losing other vital information.

\textbf{Phase \#2: cross-modal transformation learning,} which aims to map visible-to-infrared (V2IR) modality differences through paired supervision.
Then the dataset consisting of about 70,000 visible-infrared image pairs is utilized for diffusion model learning the mapping relationship between visible and infrared images. 
Subsequent to this stage, the model can generate images well consistent with the characteristics of infrared modality under the guidance of corresponding visible images. 
As the style of infrared images is relevant to many factors such as wavelength range and infrared camera sensors, the infrared images in our collected dataset have a high diversity.
This diversity strengthens the generalization capability of our model when facing all kinds of visible images, which enables the model to serve as a pre-trained model for visible-to-infrared image translation.

\textbf{Phase \#3: stylization refinement,}  which aims to adapt infrared outputs to spatio-temporal variations and environmental dynamics.
Although the pre-trained model is now capable of translating visible image into high-quality infrared one, the diversity of the training dataset makes it hard to generate infrared images in a specific style. To compensate for this shortcoming, we introduce the last training phase of our proposed progressive learning using a small dataset containing image pairs of visible images and infrared images in the desired style. 

The diffusion model gradually advances through three progressive stages of enhanced learning, ultimately evolving into a model proficient in style-controllable transformation from visible images to infrared images.

\subsection{Semantic-aware V2IR Translation via Vision-Language Understanding}
\label{sec:semantics}

V2IR is highly dependent on semantic information. 
Infrared imaging captures and visualizes the infrared radiation emitted by objects on the basis of their temperature and radiant existance.
Therefore, V2IR relies on scene understanding such as object semantics and context information (e.g., solar radiation, lighting, and shadow). 
We integrate VLUM into a unified optimization framework and thus achieve semantic-aware V2IR translation. What's more, we also incorporate additional embeddings of the segmentation map for better sturcture preserving.


To enhance the content awareness of the translation process, we use Blip~\cite{li2022blip} to create vision-language embeddings derived from visible images. This vision-language model provides a detailed description of key objects that determine the existence of radiants, along with contextual information such as weather, lighting, and other environmental elements that influence temperature. We employ a similar method to SD for text conditioning, utilizing a CLIP-based text encoder~\cite{clip} that takes text as input and applies a cross-attention mechanism to incorporate the encoded tokens. Thanks to the robust text-image alignment capability of the pre-trained stable diffusion model and the internalization of infrared representation via the progressive learning module, vision-language capabilities enable DiffV2IR to comprehend the correspondences and distinctions in cross-modality images more effectively. In addition, to maintain structural integrity in the translation process, we add embeddings from the segmentation map generated by SAM~\cite{sam}, which have a rich knowledge of layout and structure. We merge the conditioning from visible images and the segmentation map by concatenating them with the noise map after latent encoding and by adding extra input channels to the first convolutional layer of the denoising U-Net. The weights of the newly introduced input channels are initialized using zero initialization.

Moreover, the Classifier-free Guidance mechanism~\cite{cfg} is utilized to enhance the controllability of generated images using conditional inputs. This technique is often used in conditional image generation to achieve a balance between sample quality and diversity. In our approach, the score network incorporates three types of conditioning: a visible image $c_{V}$, a segmentation map $c_{S}$, and a vision-language $c_{T}$. 
During training, certain conditionings are randomly set to none to allow unconditional training, with 2\% of examples varying from fully unconditioned to having only one conditioning. To balance the control strength of the three conditionings, we introduce the following guide scales: $s_{V}$ for the visible image, $s_{S}$ for the segmentation map, and $s_{T}$ for vision-language. 
The score estimation is formulated as in Eq.~\ref{eq:2}. Each conditioning is assigned a guidance scale to adjust its intensity, resulting in a score estimate that combines conditional and unconditional outputs with specific weights.
\begin{equation}
\begin{split}
\tilde{\epsilon}_{\theta}(z_t, c_V,c_S, c_T) = \epsilon_{\theta}(z_t, \varnothing, \varnothing, \varnothing) \quad\quad\quad\quad\quad\quad\quad\quad\quad\\
+ s_V \cdot (\epsilon_{\theta}(z_t, c_V, \varnothing, \varnothing) -  \epsilon_{\theta}(z_t, \varnothing, \varnothing, \varnothing)) \quad\quad\\
+ s_S \cdot (\epsilon_{\theta}(z_t, c_V, c_S, \varnothing) - \epsilon_{\theta}(z_t, c_V, \varnothing, \varnothing))\quad\\
+ s_T \cdot (\epsilon_{\theta}(z_t, c_V, c_S, c_T) - \epsilon_{\theta}(z_t, c_V,  c_S, \varnothing))
\end{split}
\label{eq:2}
\end{equation}


\section{IR-500K Dataset}
\label{sec:dataset}
To achieve superior quality in visible-to-infrared image translation, this research brings together nearly all large publicly available datasets~\cite{kaist,FLIR,iray,jia2021llvip,Liu_2022_CVPR,Zhang_CVPR22_VTUAV,Tang2022PIAFusion,liu2020lsotb,han2023aerial,sun2020drone}, supplemented by some additional data sourced online. 
The result is an extensive multi-wavelength database comprising 500,000 infrared images. 
These images represent a wide range of scene types, diverse object categories, and various camera perspectives, such as natural landscapes, cityscapes, driving environments, aerial scene understanding, and surveillance contexts. Each image is captured at a high resolution, showcasing rich visual details, and is carefully chosen to meet the model's learning needs.
These images serve as essential components in the understanding of infrared imaging, accompanied by the label text description ``\textit{An infrared image}''.

Additionally, we selected 70,000 visible and infrared image pairs specifically for precise training purposes, as these pairs are crucial for accurately capturing the differences between the two spectra. 
These paired datasets undergo strict alignment and segmentation processing to ensure that multi-modal information can be co-learned effectively, enhancing the cross-spectral translation performance. By integrating multiple source datasets, we not only expand the training data volume but also optimize data diversity and relevance for the diffusion model, providing abundant and representative learning materials. 
With these high-quality large-scale datasets, we believe that diffusion models can effectively capture cross-spectral visual feature relationships, enabling efficient and accurate visible-to-infrared image translation tasks. The processing details of creating the IR-500K dataset will be available at \url{https://github.com/LidongWang-26/DiffV2IR}.


\section{Experiments}
\subsection{Experimental Settings}
\textbf{Testing Dataset.} 
All of our experimental evaluations are performed using the M$^3$FD dataset~\cite{Liu_2022_CVPR} and FLIR-aligned dataset~\cite{zhang2020multispectral}, both of which offer a rich array of scenes characterized by diverse weather and lighting conditions. The M$^3$FD dataset includes a total of 4,200 precisely aligned infrared and visible image pairs, each with dimensions of 1024 × 768, spread across over 10 different sub-scenarios. It is important to note that images from the same scene tend to be similar, which poses a risk of data leakage if these images are randomly splited as training and testing data.
To address this, we opt for a manual split of the M$^3$FD dataset, creating a training set composed of 3,550 pairs and a testing set containing 650 pairs, rather than relying on a random division.
The differences can be seen as shown in Table~\ref{tab:dataleak}. 
The FLIR-aligned dataset, derived from the original FLIR dataset~\cite{FLIR}, is captured from a driving perspective and has been meticulously aligned~\cite{zhang2020multispectral}. It consists of 5,142 image pairs. However, because of differences in the receptive fields between visible and infrared images, some visible images exhibit noticeable black borders. To improve the quality of the dataset, we refined to select 4,489 image pairs and divided them into training (80\%) and testing (20\%) subsets. Each image maintains a resolution of 640 × 512.


\begin{table}[htbp]
\centering
\caption{Data leak affects model performance. Experiments conducted on the identical M$^3$FD dataset but with varying segmentations exhibit significant discrepancies, suggesting that randomly dividing the M$^3$FD dataset might lead to data leakage.}
\begin{tabular}{cccc}
\toprule
\multicolumn{1}{c|}{Dataset} & FID↓ & PSNR↑ &SSID↑   \\ \cline{1-4}
\multicolumn{1}{c|}{M$^3$FD (randomly split)} &  45.89 & 21.70 & 0.7196 \\
\multicolumn{1}{c|}{M$^3$FD (manually split)} & 70.29 & 19.30 & 0.6620\\
\bottomrule
\end{tabular}
\label{tab:dataleak}
\end{table}

\textbf{Visual Quality Assessment.} 
We assess the quality of translated images using established standard metrics, such as Fr\'echet Inception Distance (FID)~\cite{NIPS2017_8a1d6947}, Peak Signal-to-Noise Ratio (PSNR)~\cite{hore2010image}, and Structural Similarity Index (SSIM)~\cite{ssim}.

\textbf{Implementation Details.} 
The experiments in this study were carried out on a system equipped with a NVIDIA A800 GPU. Throughout training, images were first resized to 286 x 286 and then randomly cropped to 256 x 256 to enhance training speed and efficiency. For models provided with a recommended configuration, the experiments were executed under those specified settings. Our DiffV2IR and models lacking recommended configurations were trained for approximately 100 epochs to ensure proper convergence. For techniques requiring text input, the same text prompt from phase \#3 of PLM training in our DiffV2IR was supplied. For style transfer methods that require both a style and content image, we randomly chose an infrared image from the training set as the style reference and used a visible image as the content to be translated.

\begin{table*}[htbp]
\centering
\setlength{\tabcolsep}{13pt}
\caption{Quantitative comparison with the state-of-the-arts. The best results are highlighted in \textbf{bold} and the second best results are \underline{underlined}. The methods in the first half of the table are GAN-based, while the latter half are based on diffusion models.}
\begin{tabular}{l|c|ccc|rrc}
\toprule
\multicolumn{1}{c|}{\multirow{2}{*}{Method}} & \multirow{2}{*}{Reference} & \multicolumn{3}{c|}{{M$^3$FD dataset}}            & \multicolumn{3}{c}{{FLIR-aligned dataset}} \\ \cline{3-8}
                  &                   & \multicolumn{1}{c}{FID↓} & \multicolumn{1}{c}{PSNR↑} & \multicolumn{1}{c|}{SSIM↑} & FID↓ &PSNR↑&SSIM↑      \\ \midrule
 Pix2Pix~\cite{pix2pix}   &   CVPR$_{17}$      & 182.14& 17.19 & 0.5672&  98.81 &\textbf{19.79} & 0.4327 \\
 CycleGAN~\cite{cyclegan}  &ICCV$_{17}$      & 114.71 & 14.98  & 0.5271 &59.74 & 16.58   & 0.4091 \\
EGGAN-U\cite{eggan,unit} &ICRA$_{23}$ & 149.12 &  13.87& 0.5455 & 113.51&  15.76  & 0.4253 \\
DR-AVIT\cite{han2024dr} &TGRS$_{24}$        & 116.96 & 14.19 & 0.5449  & 65.96 & 16.30 & 0.4355 \\ 
StegoGAN\cite{Wu_2024_CVPR}   &CVPR$_{24}$  & 183.56  & 13.19  & 0.4303 & 87.57  & 12.44  & 0.3752\\
UNSB\cite{UNSB}       &ICLR$_{24}$           & 115.94  & 14.07  & 0.4885 & 85.61  & 9.95  & 0.3179\\\hline
ControlNet\cite{controlnet}     &ICCV$_{23}$ & 140.14 & 15.17 & 0.5572  & 119.69 & 11.98 & 0.2783 \\ 
FCDiffusion\cite{fcdiffusion}  &AAAI$_{24}$  & 170.14 & 11.60 & 0.2854 & 180.58 & 10.89 & 0.2860\\ 
T2I-Adapter\cite{t2i}     &AAAI$_{24}$        &114.63 &15.98  & 0.5976 & 91.61 & 12.33  & 0.3689\\ 
OSASIS\cite{Cho_2024_CVPR}  &CVPR$_{24}$      & 243.21  & 14.59 & 0.5642 & 192.44 & 14.51  & 0.3774 \\
StyleID\cite{styleid}     &CVPR$_{24}$    & 135.97 & 12.67 & 0.4317 & 94.28  & 10.69  & 0.3086\\ 
CSGO\cite{xing2024csgo}     &ARXIV$_{24}$        & 185.32  & 10.33 & 0.4147 & 178.04  & 9.63  & 0.3288                                    \\
Pix2PixTurbo\cite{parmar2024one}  &ARXIV$_{24}$   & \multicolumn{1}{r}{98.12} & 16.80 & 0.5964 & 90.72  & 15.92  & \underline{0.4590} \\
PID\cite{mao2024pid}     &ARXIV$_{24}$        & 160.91 & 16.10 & 0.5579 & \underline{43.98} & \underline{18.89} & 0.4315\\ 
InstructPix2Pix\cite{ip2p} &CVPR$_{23}$ & \multicolumn{1}{r}{\underline{81.64}} & \underline{17.92}  & \underline{0.6328} & 46.29 & 18.41 & 0.4481 \\ \hline
\textbf{DiffV2IR}      & \textbf{Ours}      & \multicolumn{1}{r}{\textbf{70.29}}  & \textbf{19.30} & \textbf{0.6620} & \textbf{39.99} & 18.63  & \textbf{0.4658}\\ 
\bottomrule  
\end{tabular}
\label{tab:sota}
\end{table*}

\subsection{Comparison with SOTA Methods}
We evaluate our DiffV2IR model compared to fifteen cutting-edge methods that have emerged in recent years, many of which necessitate additional training before deployment. These methods include GAN-based methods such as Pix2Pix~\cite{pix2pix}, CycleGAN~\cite{cyclegan}, EGGAN-U~\cite{eggan,unit}, DR-AVIT\cite{han2024dr}, StegoGAN~\cite{Wu_2024_CVPR}, UNSB~\cite{UNSB} and methods based on diffusion models like InstructPix2Pix~\cite{ip2p}, ControlNet~\cite{controlnet}, FCDiffusion~\cite{fcdiffusion}, T2I-Adapter~\cite{t2i}, Pix2PixTurbo~\cite{parmar2024one}, PID~\cite{mao2024pid}, among others. In addition, we also examine pre-trained models (CSGO~\cite{xing2024csgo}), training-free approaches (StyleID~\cite{styleid}), and few-shot methods (OSASIS~\cite{Cho_2024_CVPR}) for comparative analysis.

\begin{figure*}[htbp]
    \centering
    \includegraphics[width=1.0\linewidth,height=4.2in]{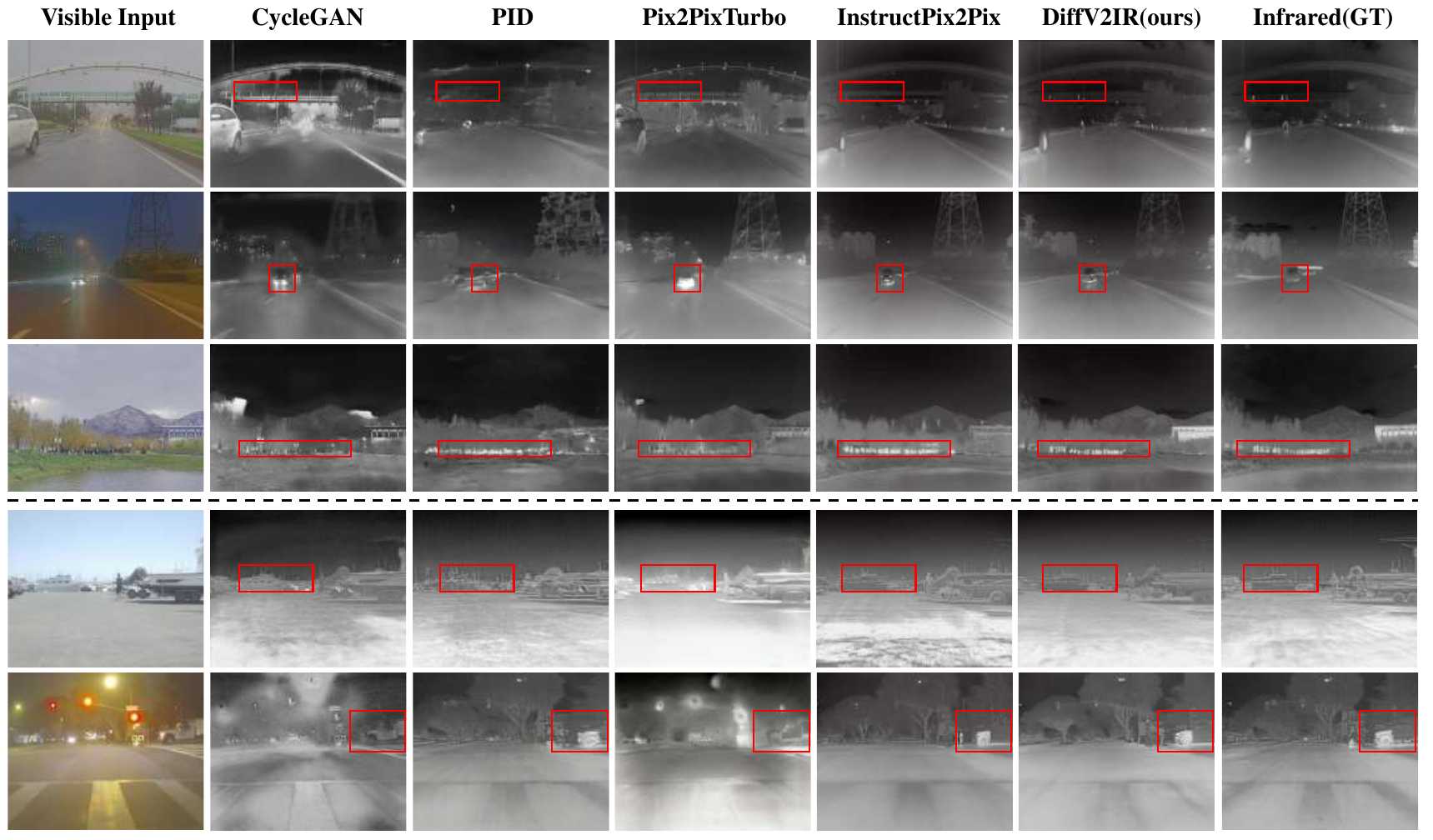}
    \caption{Comparison with SOTA methods. Key differences are highlighted within a red box, such as halos and low-light scenarios. Only the top 5 methods according to assessment metrics are shown. (Top: results from M$^3$FD dataset; Bottom: results from FLIR-aligned dataset.)}
    \label{fig:sota}
\end{figure*}


\textbf{Quantitative Comparisons.} 
Table~\ref{tab:sota} provides information on the performance of V2IR translations.

On the M$^3$FD dataset, Pix2Pix~\cite{pix2pix} stands out among GAN-based approaches with the highest PSNR and SSIM, signifying excellent pixel-level precision, though it also records one of the worst FID scores. CycleGAN~\cite{cyclegan}, DR-AVIT~\cite{han2024dr}, and UNSB~\cite{UNSB} offer a more balanced performance, yet their results remain unsatisfactory. Although EGGAN-U~\cite{eggan,unit} achieves the second-best SSIM among GAN methods, its overall performance along with StegoGAN~\cite{Wu_2024_CVPR} is not commendable. Regarding diffusion models, ControlNet~\cite{controlnet} and T2I-Adapter~\cite{t2i} achieve impressive PSNR and SSIM scores by transforming visible images into feature maps for conditional guidance in the denoising process, with T2I-Adapter~\cite{t2i} reaching a low FID of 114.63. On the other hand, FCDiffusion~\cite{fcdiffusion} struggles with input visible image structure preservation, showing the worst SSIM. Methods like OSASIS~\cite{Cho_2024_CVPR}, training-free approaches such as StyleID~\cite{styleid}, and the pre-trained CSGO model~\cite{xing2024csgo} fail to deliver good metrics due to insufficient training data and limited infrared understanding. PID~\cite{mao2024pid}, another diffusion model for infrared generation, presents reasonable PSNR and SSIM but suffers from a high FID. In contrast, InstructPix2Pix~\cite{ip2p} and Pix2PixTurbo~\cite{parmar2024one} excel with the second and third best metrics across all methods.

In contrast to the M$^3$FD dataset, the FLIR-aligned dataset presents a less complex scenario. GAN-based methods demonstrate improved outcomes over their performance on the M$^3$FD dataset, with Pix2Pix~\cite{pix2pix} achieving the highest PSNR among all methods. Nevertheless, the FID score for Pix2Pix~\cite{pix2pix} remains significantly elevated. Similar to their results on the M$^3$FD dataset, CycleGAN~\cite{cyclegan} and DR-AVIT~\cite{han2024dr} maintain balance across three metrics, whereas EGGAN-U~\cite{eggan,unit}, StegoGAN~\cite{Wu_2024_CVPR}, and UNSB~\cite{UNSB} continue to exhibit subpar performance. As for diffusion-based models, ControlNet~\cite{controlnet}, T2I-Adapter~\cite{t2i}, and FCDiffusion~\cite{fcdiffusion} still struggle to adjust effectively to the V2IR task. Methods lacking additional training also show weak performance. Pix2PixTurbo~\cite{parmar2024one} ranks second best in SSIM, although its FID remains high. PID~\cite{mao2024pid} performs notably well, achieving the second-best FID and PSNR. InstructPix2Pix~\cite{ip2p} delivers a well-rounded performance with balanced metrics.

Based on experimental outcomes and previous observations, models utilizing GANs tend to excel with simpler datasets as opposed to more intricate scenes. Furthermore, several techniques are significantly affected by mode collapse, necessitating multiple training rounds to guarantee high-quality generation. InstructPix2Pix~\cite{ip2p}, Pix2PixTurbo~\cite{parmar2024one}, and our proposed DiffV2IR avoid converting the visible image into a noise or feature map, which we believe contributes to their superior performance compared to other methods.

\textbf{Qualitative Comparisons.} As illustrated in Figure~\ref{fig:sota}, the proposed DiffV2IR offers significant improvements in both global image quality and control over local details. Current methods face several key issues, the most critical being incorrect handling of halos around light sources, primarily due to a lack of understanding of the infrared modality. In visible images, halo regions often exhibit high brightness, whereas in infrared images, only objects emitting heat should appear brighter. Another issue arises when processing visible images in low-light conditions, where some methods fail to maintain the image structure and details effectively. In contrast, DiffV2IR leverages enhanced infrared knowledge and vision-language understanding to produce credible images that adhere to physical principles while providing outstanding detail management.

The experiments demonstrate the effectiveness of diffusion models in multi-spectral translation tasks and validate the robustness of our proposed DiffV2IR under different conditions.

\subsection{Ablation Study}
We perform an ablation study on the M$^3$FD dataset. Table~\ref{tab:ablation} highlights the impact of our progressive learning and vision-language understanding modules.

\begin{table}[!b]
\centering
\caption{The ablation study is split into two distinct stages. The initial rows focus on evaluating the training phases of progressive learning excluding VLUM, whereas the later rows assess the types of information in VLUM when combined with PLM.}
\resizebox{0.5\textwidth}{!}{
\begin{tabular}{ccc|cc|ccc}
\toprule 
\multicolumn{3}{c|}{PLM} & \multicolumn{2}{c|}{VLUM} & \multirow{3}{*}{FID↓} & \multirow{3}{*}{PSNR↑} & \multirow{3}{*}{SSIM↑} \\ \cline{1-5}
{\begin{tabular}[c]{@{}c@{}} \#1\end{tabular}} &
  {\begin{tabular}[c]{@{}c@{}}\#2\end{tabular}} &
  {\begin{tabular}[c]{@{}c@{}}\#3\end{tabular}} &
  {\begin{tabular}[c]{@{}c@{}} Seg.\\ Map\end{tabular}} &
  {\begin{tabular}[c]{@{}c@{}} Vision-\\ Language\end{tabular}} &
   &
   &
   \\
  \midrule
 
×                              & \checkmark                               & ×                              & -                               & -                              & 113.98 & 12.71 & 0.4013 \\
 
\checkmark                              & \checkmark                             & ×                              & -                               & -                              & 112.45 & 13.50 & 0.4055 \\

×                              & ×                               & \checkmark                              &-                              & -                              & 81.15  & 18.47 & 0.6439 \\
\checkmark                              & ×                               & \checkmark                              &-                              &-                              & 78.10  & 18.71 & 0.6481 \\
×                              & \checkmark                               & \checkmark                              &-                              &-                              & 75.48  & 19.01 & 0.6533 \\
\checkmark                              & \checkmark                               & \checkmark                              &-                           &-                              & 74.79  & 19.13 & 0.6557 \\
\hline 
\hline
\checkmark                              & \checkmark                               & \checkmark                              & ×                               & \checkmark                              & 73.92  & 19.11 & 0.6563 \\
 
\checkmark                              & \checkmark                               & \checkmark                              & \checkmark                               & ×                              & \underline{71.63}  & \underline{19.17} & \underline{0.6585} \\
 
\checkmark                              & \checkmark                               & \checkmark                              & \checkmark                               & \checkmark                              & \textbf{70.29}  & \textbf{19.30} & \textbf{0.6620} \\ 
\bottomrule
\end{tabular}
}
\label{tab:ablation}
\end{table}

\textbf{Progressive Learning Module.} We conduct experiments using all possible combinations of the three phases, relying exclusively on visible-infrared image pairs without any additional embeddings. Since the original Stable Diffusion functions as a text-to-image framework and models fine-tuned only in phase \#1 are unable to perform image translation, we excluded that case. Each training phase contributes positively to the results.

\textbf{Vision-Language Understanding Module.} After establishing the progressive learning approach, we initially incorporated two distinct embeddings for vision-language and segmentation maps independently and subsequently combined them. Although each additional individual embedding enhances the quality of generation, the unified strategy, known as the DiffV2IR method, achieves the highest level of performance.

\textbf{Hyper-parameters.} A limited number of hyperparameters can influence the ultimate performance. Primarily, we experiment with denoising steps and classifier-free guidance scales across three conditional inputs. Table~\ref{tab:hyper} presents the results. 
The ultimate outcome is a balance among several factors, such as inference consumption and translation quality according to these three evaluation metrics.

\begin{table}[]
\centering
\caption{Ablation study of denoising steps and classifier-free guidance scales. $s_{T}$, $s_{V}$, $s_{S}$ denote guidance scale for vision-language, visible image and segmentation mask, respectively. The optimal settings we choose are highlighted in bold. }
\setlength{\tabcolsep}{10pt}
\begin{tabular}{c|c|ccc}
\toprule
          \multicolumn{2}{c|}{Hyper-parameters}                           &  FID↓ & PSNR↑ & SSIM↑ \\ \hline
\multirow{4}{*}{{steps}}      & 50        & 70.92      & 19.37       & 0.6640        \\
                                     & \textbf{100}       & \textbf{70.29}      & \textbf{19.30}       & \textbf{0.6620}        \\
                                     & 150       & 71.41      & 19.38       & 0.6616        \\
                                     & 200       & 70.69      & 19.33       & 0.6614        \\ \hline
\multirow{3}{*}{\textbf{$s_{T}$}}        & 5.0       & 71.34      & 19.43       & 0.6656        \\
                                     & \textbf{7.5}       & \textbf{70.29}      & \textbf{19.30}       & \textbf{0.6620}        \\
                                     & 10.0      & 72.04      & 19.34       & 0.6592        \\ \hline
\multirow{3}{*}{\textbf{$s_{V}$}}        & 1.0       & 71.65      & 19.34       & 0.6663        \\
                                     & \textbf{1.5}       & \textbf{70.29}      & \textbf{19.30}       & \textbf{0.6620}        \\
                                     & 2.0       & 73.02      & 19.24       & 0.6580        \\ \hline
\multirow{3}{*}{\textbf{$s_{S}$}}        & 1.0       & 72.09      & 19.32       & 0.6630        \\
                                     & \textbf{1.5}       & \textbf{70.29}      & \textbf{19.30}       & \textbf{0.6620}        \\
                                     & 2.0       & 72.45      & 19.37       & 0.6631        \\
\bottomrule
\end{tabular}
\label{tab:hyper}
\end{table}

\section{Conclusion}
Converting visible images into infrared images is highly demanded and is not adequately addressed. The primary challenges are generating content with semantic awareness, differences in spectrum appearances, and the scarcity of effective infrared datasets.
This study introduces DiffV2IR, an innovative framework for translating visible images into infrared images. 
By integrating PLM, VLUM and the comprehensive IR-500K dataset, we significantly enhance the V2IR translation performance. Experimental findings confirm the effectiveness of diffusion models in generating superior translations, demonstrating their efficacy and wide-ranging applicability, thereby offering a fresh approach for multi-spectral image generation. 

\noindent\textbf{Limitation.} DiffV2IR is specialized in translating broad scenes, which may restrict its effectiveness in specific applications such as face image translation.

\noindent\textbf{Potential Negative lmpact.} DiffV2IR focuses on infrared image synthesis. It might be misused to create misleading content.

{
    \small
    \bibliographystyle{ieeenat_fullname}
    \bibliography{main}
}

\maketitlesupplementary
\setcounter{section}{0} 
\renewcommand{\thetable}{\Roman{table}}
\renewcommand{\thefigure}{\Roman{figure}}

\setcounter{figure}{0}
\setcounter{table}{0}
\renewcommand{\thesection}{{\Alph{section}}}
\setcounter{page}{1}

To provide more details of our proposed DiffV2IR, this supplementary material includes the following content:
\begin{itemize}[leftmargin=15pt]
    \item Section \ref{app:IR-500K}: Common infrared image datasets that have been widely used in recent years. 
    \item Section \ref{app:intermediate}: Intermediate results for a vivid understanding of our proposed modules.
    \item Section \ref{app:visual}: Infrared images generated by all the SOTA methods mentioned in our paper.
    \item Section \ref{app:video}: A demo video to better illustrate our motivation and solving approach.
\end{itemize}

\section{Comparison of Our Proposed IR-500K and Other Commonly Used Infrared Datasets}
\label{app:IR-500K}

Table~\ref{tab:data} provides a comparison of our proposed IR-500K and other commonly used infrared datasets. Though the scale is not the biggest, IR-500K has the best diversity with various scenarios, multiple camera angles, and all kinds of object. The scenarios of IR-500K include urban scenes, such as campus, parks, roads, infrastructures, and natural scenes such as rivers, lakes, beaches, seas, mountains. The camera angles consist of aerial view by drones, surveillance view by monitors, driving view by in-vihecle cameras, horizontal view by handheld cameras, and so on. The main objects comprise human beings, vehicles, wild animals, natural landscapes, and buildings.

\begin{table*}[!b]
\centering
\caption{Comparison of our proposed IR-500K and other commonly used visible-infrared datasets.}
\begin{tabular}{lccc}
\toprule
Dataset      & Camera Angle & Scenario                  & Amount                                       \\
\midrule
MSRS~\cite{Tang2022PIAFusion}         & Driving      & Road                                       & 1,444 pairs                                    \\
AVIID~\cite{han2023aerial}        & Aerial       & Road                                     & 3,363 pairs                                    \\
M$^3$FD~\cite{Liu_2022_CVPR}         & Horizontal   & Campus, Road, Natural scenes              & 4,200 pairs                                    \\
FLIR~\cite{FLIR}         & Driving      & Road                                  & 9,711 IR/9,233 visible (not aligned) \\
LLVIP~\cite{jia2021llvip}        & Surveillance & Street                                & 15,488 pairs                                   \\
DroneVehicle~\cite{sun2020drone} & Aerial       & Road, Urban area                   & 28,439 pairs                                   \\
Kaist~\cite{kaist}        & Driving      & Road                                & 95,000 pairs                                    \\
VTUAV~\cite{Zhang_CVPR22_VTUAV}        & Aerial       & Urban scenes                        & 1.7M pairs from 500 sequences                \\ \hline
IR-500K(\textbf{ours})      & Multiple     & Multiple                             & 500K IR, 70,000 pairs            \\ 
\bottomrule
\end{tabular}
\label{tab:data}
\end{table*}

\section{Intermediate Results}
\label{app:intermediate}
\begin{figure*}[!hbp]
    \centering
    \includegraphics[width=1\linewidth]{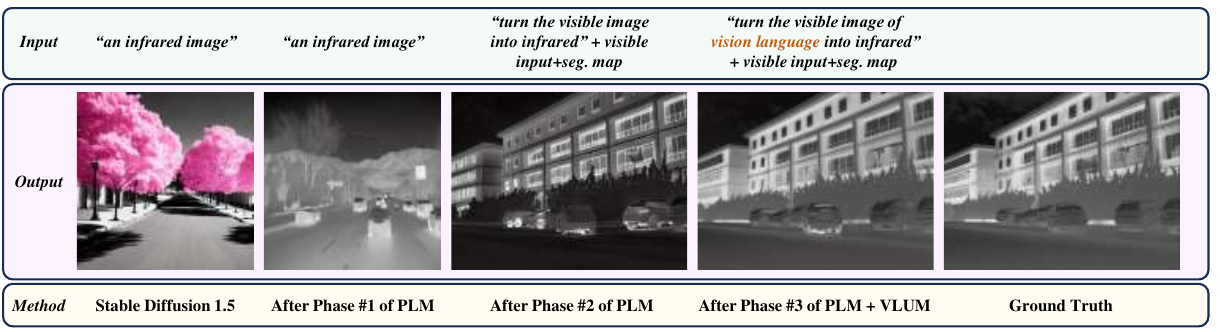}
    \caption{Intermediate results of our proposed DiffV2IR.}
    \label{fig:supp3}
\end{figure*}

Figure~\ref{fig:supp3} presents intermediate outcomes of PLM and VLUM, highlighting their effectiveness. Although original stable diffusion (v 1.5) excels at generating images from text owing to extensive training data, it struggles with the infrared modality due to limited infrared knowledge. Hence, the generated output is far from a normal infrared image. 
In Phase \#1 of PLM, the model internalizes the infrared representation, allowing it to produce high-quality infrared images, although without precise control. 
From Figure~\ref{fig:supp3} column 2, the output of PLM Phase \#1 shows good appearance as an infrared image. 
During Phase \#2, the model starts to convert visible images into infrared ones with a guidance visible imange and a segmantation map, maintaining structural integrity, though stylization remains challenging due to the tight link between semantics and infrared imagery. 
Finally, with the refinement of stylization in phase \#3 and the incorporation of VLUM, the model achieves both structure preservation and semantic awareness.
Our final outputs of DiffV2IR can have both good texture matching and structural preservation.

\section{More Visualization Results}
\label{app:visual}
We present the visual results of infrared images generated by 15 state-of-the-art (SOTA) methods involved in the experimental comparison section. 

\begin{figure*}[htbp]
    \centering
    \includegraphics[width=0.91\linewidth]{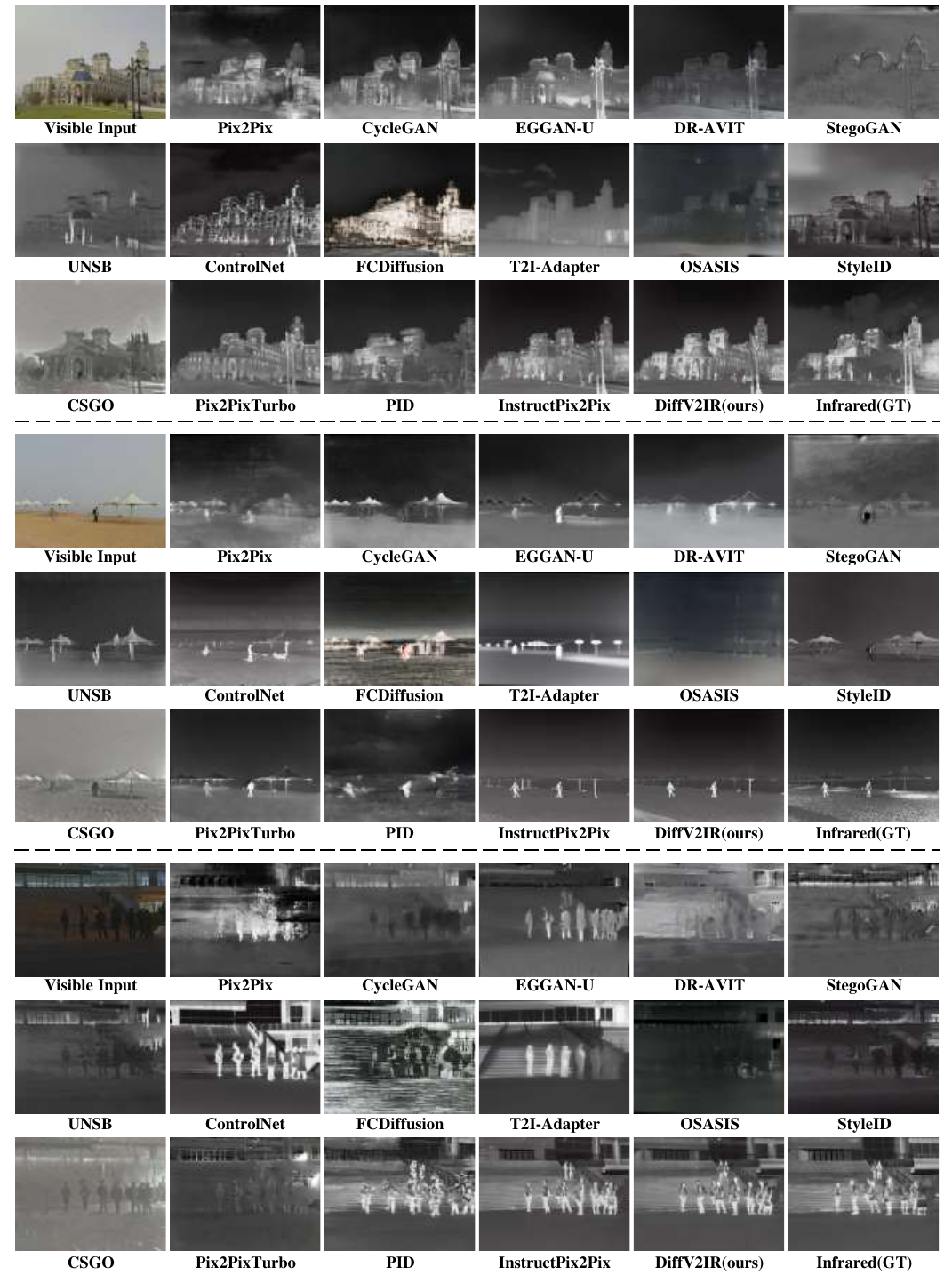}
    \caption{Comparison with SOTA methods on M$^3$FD dataset.}
    \label{fig:supp1}
\end{figure*}

\begin{figure*}[htbp]
    \centering
    \includegraphics[width=0.9\linewidth]{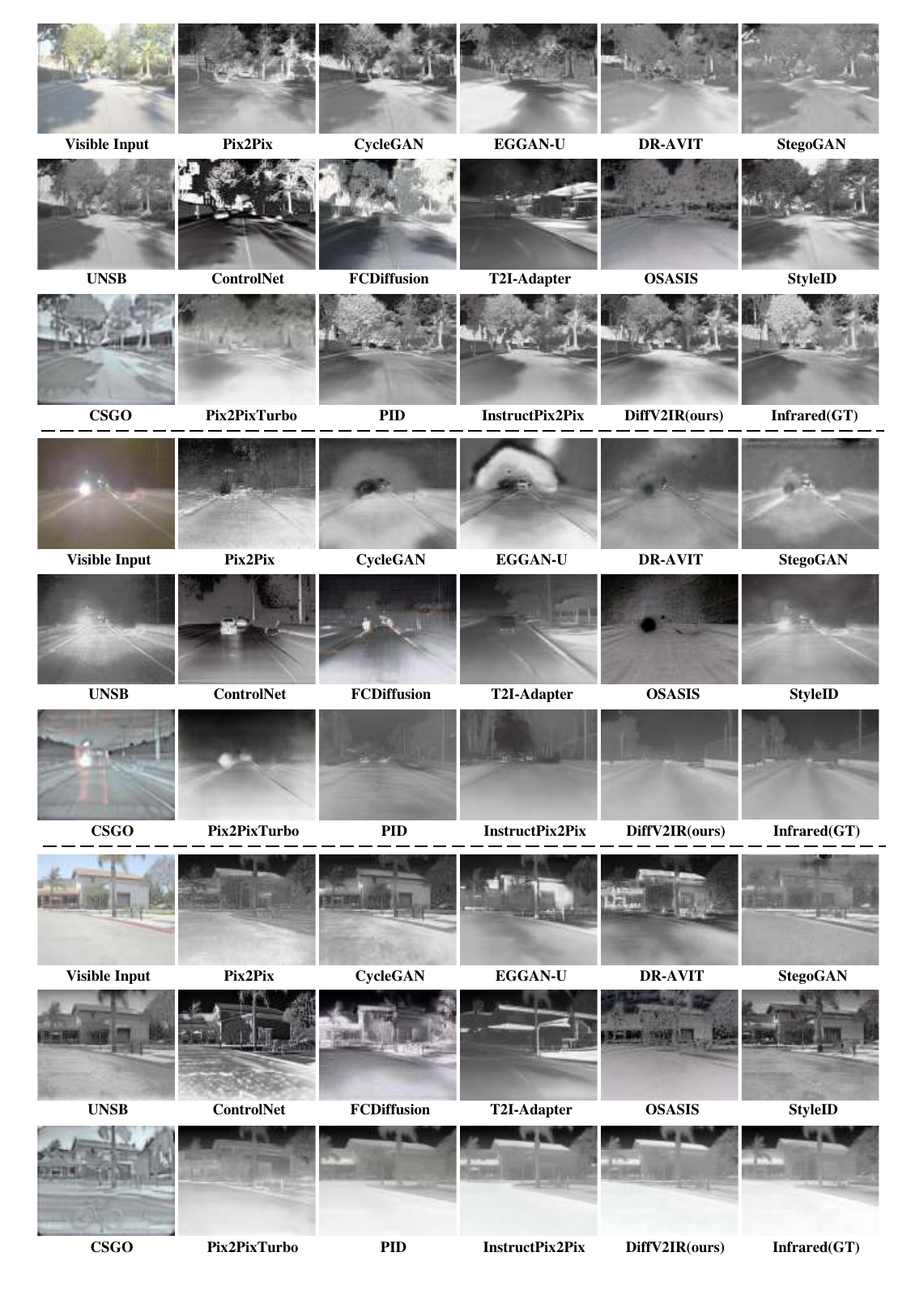}
    \caption{Comparison with SOTA methods on FLIR-aligned dataset.}
    \label{fig:supp2}
\end{figure*}

Figure~\ref{fig:supp1} and Figure~\ref{fig:supp2} show the visualization results of all the methods mentioned in our experimental part, on M$^3$FD~\cite{Liu_2022_CVPR} and FLIR-aligned~\cite{zhang2020multispectral} dataset. There are 
several main problems. The first one is that some models like Pix2Pix~\cite{pix2pix} and PID~\cite{mao2024pid} exhibit ambiguity, which causes Image quality degradation. The second one is the failure of preserving structure of visible inputs, such as StyleID~\cite{styleid} and CSGO~\cite{xing2024csgo}. The third as well as the most common problem is the overlook of semantics and context information.

\section{Demo Video}
\label{app:video}
To effectively convey the research's objectives, challenges, and innovative aspects, we have created a video that outlines three principal challenges in infrared image translation, describes the technical methodology for converting visible to infrared images employed in this study, and highlights preliminary implementation results.

Please refer to the demo video attached in ``supplementary.zip".

\end{document}